\documentclass[runningheads]{llncs}
\usepackage{graphicx}
\usepackage{amsfonts}
\usepackage[english]{babel}
\usepackage[utf8]{inputenc}
\usepackage{algorithm}
\usepackage[noend]{algpseudocode}
\usepackage{diagbox}
\usepackage{color}
\usepackage{url} 
\usepackage{breakurl} 
\usepackage{hyperref} 
\usepackage{amsmath}
\usepackage[margin=1in]{geometry}

\begin{document}
\title{Assessing the trade-off between prediction accuracy and interpretability for topic modeling on energetic materials corpora}
%
%
\author{Monica Puerto\inst{1} \and
Mason Kellett\inst{1} \and
Rodanthi Nikopoulou\inst{1} \and
Mark D. Fuge\inst{2} \and
Ruth Doherty\inst{3} \and
Peter W. Chung\inst{2} \and
Zois Boukouvalas\inst{1}}
\authorrunning{Puerto et al.}
%
\institute{American University, Washington, D.C. 20016, USA\\
\and
Center for Engineering Concepts Development, Department of Mechanical Engineering, University of Maryland, College Park, MD\\
\and
Energetics Technology Center, Indian Head, MD}
\maketitle              
\begin{abstract}
As the amount and variety of energetics research increases, machine aware topic identification is necessary to streamline future research pipelines. The makeup of an automatic topic identification process consists of creating document representations and performing classification. However, the implementation of these processes on energetics research imposes new challenges. Energetics datasets contain many scientific terms that are necessary to understand the context of a document but may require more complex document representations. Secondly, the predictions from classification must be understandable and trusted by the chemists within the pipeline. In this work, we study the trade-off between prediction accuracy and interpretability by implementing three document embedding methods that vary in computational complexity. With our accuracy results, we also introduce local interpretability model-agnostic explanations (LIME) of each prediction to provide a localized understanding of each prediction and to validate classifier decisions with our team of energetics experts. This study was carried out on a novel labeled energetics dataset created and validated by our team of energetics experts.

\keywords{Energetics; machine learning; natural language processing; topic modeling; interpretability.}
\end{abstract}
\section{Introduction}
With the increasing amount of energetics research available, the time spent searching for relevant research documents can be inefficient. Machine-aware topic identification can streamline this process by identifying relevant research efficiently. Topic identification can be broken down into three common steps. First, the creation of word or document embeddings that transform text rich documents into vector representations of each document that hold latent semantic information about the context within each document. These document embeddings are the input to a classification algorithm where a machine learning classifier is trained on pre-labeled documents and tested on unseen unlabeled documents. The final step is to evaluate which classifier is most accurate using common machine learning evaluation metrics \cite{japkowicz2011evaluating}.

This process is a common natural language processing (NLP) pipeline for classification or topic identification problems; however, the implementation of this pipeline for energetics research introduces new challenges. The documents contain many different chemicals and ways to represent these chemicals, so one must identify and deal with these accordingly \cite{elton2019using}. Secondly, there are many different topics that exist within and around the field of energetics. Frequently, one energetics research paper will have more than one relevant topic within its context. Therefore, to effectively address those challenges, in this work we designed the experiments so we could identify which embedding technique would best identify and differentiate between topics. This leads to the final challenge, which in the area of machine learning is referred as machine learning interpretability \cite{linardatos2021explainable,hansen2019interpretability}. As a researcher, if you cannot identify why a research paper was predicted to have a certain label, then a topic identification system has little significance. Therefore, to address the last challenge we implement local interpretable model-agnostic (LIME) explanations \cite{ribeiro2016should}, for each topic identification pipeline to weigh against traditional evaluation metrics.

In this paper, we introduce an experiment to identify which of Latent Dirichlet Allocation (LDA) \cite{blei2003latent}, Word2vec \cite{mikolov2013efficient}, and Bidirectional Encoder Representations (BERT) \cite{BERT} embedding techniques most efficiently and logically classify energetics documents. LDA is a probabilistic statistical method that estimates the prior probabilities for a specific document to align with a certain topic for any given number of topics. Word2vec is a shallow neural network where the hidden layer creates a weighted matrix that is used for the word embeddings. BERT is a deep neural network that has been pre-trained with Google AI. All three techniques can be used to perform the same task, i.e., generate document embeddings/representations, but differ in algorithmic and computational complexity. We will then demonstrate how they compare to one another using traditional evaluation metrics and LIME in order to understand the accuracy and interpretability trade-off we experience. The paper concludes with recommendations for future research and applications of NLP in the field of energetics.

The remainder of the paper is organized as follows. In section \ref{MaterialsMethods}, we discuss our materials and methods which contain our NLP pipeline, the development of our energetics dataset, the machine learning framework, as well as how we implement local interpretability to our pipeline. The experimental set-up and our results are presented in Section \ref{results}. Section \ref{conclusion}, concludes the paper.

\section{Materials and methods}\label{MaterialsMethods}

\begin{figure}[H]
\centering\includegraphics[width=13cm]{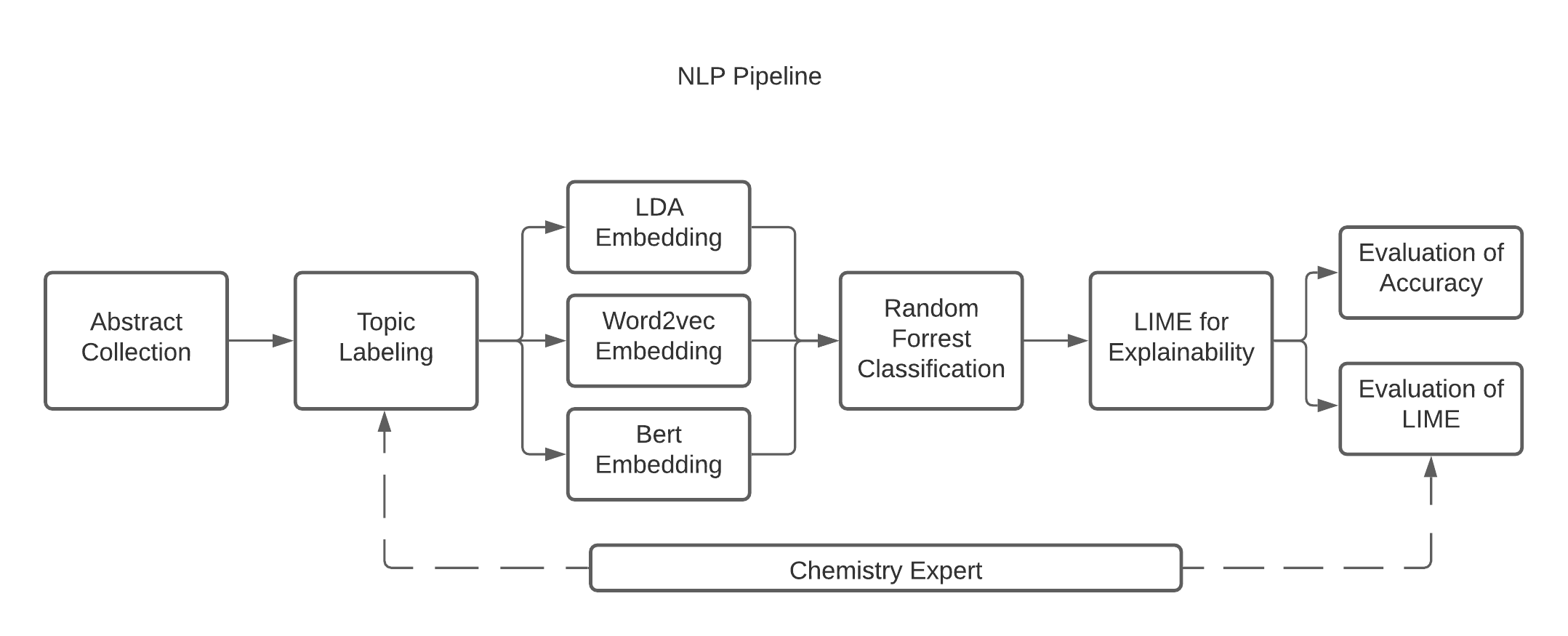}
\centering
\caption{Energetic materials abstract classification pipeline}
\label{pipeline}
\centering
\end{figure}

The NLP pipeline used in this work can been seen in Figure \ref{pipeline}. In this section we present the main components of our NLP pipeline; the abstract collection done by our chemistry experts, the data labeling done by the same chemistry experts, the creation of three distinct document embedding datasets, fitting the classifier with these embeddings individually, and assessing the LIME outputs. It is worth noting that the implementation of our pipeline was constructed in Python 3.8\footnote{Python Software Foundation. Python Language Reference. Available at \url{http://www.python.org}} by utilizing many python packages such as scikit-learn\footnote{\url{https://scikit-learn.org/stable/}} for the classifier, LIME for explainability, transformers for the BERT model, and the Gensim package\footnote{\url{https://pypi.org/project/gensim/}} for the Word2vec model and LDA model.

The energetics abstract were text files provided by our chemistry experts, which two of them then labeled the abstracts into multiple topics which is further explained in subsection 2.1. With the labeled data we used three individual models to create different document embeddings which is further explained in subsection 2.2. We then fit these matrices representing embeddings individually in a random forest model. After each model is fit we extrapolate the LIME observations using the LIME package with the fitted model for each testing observation. Finally we evaluate not just the prediction performance of the model but also the level of agreement with the LIME output and a chemistry expert that aided in the manual labeling.

\subsection{Development of an energetics dataset}
To develop our energetics dataset we performed a double blind study, where two chemists ranging in experience, Dr. Ruth Doherty and Rodanthi Nikopoulou, were separately given energetic materials scientific abstracts for labeling. Initially, Dr. Ruth Doherty characterized 700 abstracts into nine distinct topic labels. Since this dataset was going to be part of a multi-label classification project and to avoid potential issues related to class imbalance \cite{japkowicz2002class}, we keep the topic labels to a maximum of four labels which encompassed 474 abstracts. The description of each label is:

\begin{itemize}
    \item[] {\it Characterization}: Relating to the physical characteristics or properties of energetic materials.
    \item[] {\it Modeling}: Relating to the methods used to model or mimic the behavior of molecules.
    \item[] {\it Processing}: Relating the modifications of energetic chemicals.
    \item[] {\it Synthesis}: Relating to the production of an energetic compound.
\end{itemize}

Next, to eliminate data bias our second subject matter expert, Rodanthi Nikopoulou, independently reviewed each abstract to determine whether or not an abstract should be labeled as characterization, modeling, processing, or synthesis. For our double blind study, the two chemists from our group defined and provided the set of rules for each of these four label assignments. The pre-defined set of rules for the topic labels are:

\begin{itemize}
    \item[(1)] The topic label or a derivative of the topic label is stated in the title of a research document.
    \item[(2)] A direct synonym of the topic label is stated in the title of the research document.
    \item[(3)] The introduction or conclusion mentions a word from a set of keywords that indicate a specific topic label.
    \item[(4)] A majority of the key words within the abstract relate to a specific topic label.
\end{itemize}

To ensure the labeling process was consistent and to eliminate human bias, we only kept the abstracts that both experts agreed on, which was 258 abstracts across these four topics out of the 474 abstracts. Note here that in order to obtain a balanced and manageable dataset our two experts also manually checked this the labels for consistency and validity with respect to reliability.

\subsection{Creating document embeddings}

Before we create our document embeddings, each document was pre-processed with lower-casing, punctuation removal and part of speech recognition to keep only nouns and verbs. The complete corpus includes the pre-processed text from the title and abstract of the document. To ensure only the pertinent information was included, we introduced bigrams with a frequency threshold.

To create the document embeddings, we converted the text rich abstracts into document embeddings and performed machine learning computations. The process of creating document embeddings is pivotal for both the accuracy of the classification algorithm and the local interpretability of an individual classification result. To determine which embedding is best suited for classification on energetics research documents, we designed an experiment to optimize three different embedding methods; LDA, Word2vec, and BERT, that increase in complexity respectively. The objective of the experiment is to understand the interaction between local interpretability and overall accuracy as the complexity of the embedding method increases. As we see in later subsection, the classification algorithm was kept constant for all iterations of the experiment.

\subsubsection{Latent Dirichlet allocation}
LDA  is a Bayesian probabilistic unsupervised topic model where the topic probabilities provide an explicit representation of a document \cite{blei2003latent}. We input the number of topics, $K$, where $K$ is tuned through cross-validation. The output of LDA contains two matrices of posterior probability distributions. One is the term to topic distributions where each word has a Dirichlet distribution of topic probabilities while the other matrix is the document to topic distribution where each document has a Dirichlet distribution of topic probabilities. High probability words identify context within the documents at a higher level. The document embeddings used for classification are the topic to term distributions from LDA. All terms will have an array of the posterior distribution belonging to the $K$ topics. After optimizing LDA using cross-validation, we improved the performance of LDA with a set of 100 topics ($K = 100$). The input to the classification algorithm is a matrix sized $N \times 100$ where $N$ is the number of abstracts.

\subsubsection{Word2Vec}
Word2Vec is a shallow neural network where the hidden layer provides hidden semantic information about the context of words within a document. The input layer is a hot encoded vector, the hidden layer is a weighted matrix that maps input to output, and the output layer contains a soft-max activation function that outputs the probability of either the center word (CBOW) or the words surrounding the center word (skip-gram) existing near one another. The hidden layer forms the word embedding for Word2Vec. The average is taken for all word embeddings from the hidden layer within a document in order to create a document embedding matrix that will be later used in a classification task. The output matrix is sized $N \times 200$, where $N$ is the number of abstracts and 200 is the number neurons in the hidden layer and was determined through cross-validation.

\subsubsection{BERT}
Bidirectional deep learning neural network built by Google AI that uses transformer architecture and attention layers for auto-encoder and decoders. BERT-Large has 24 layers with an embedding size of 1024 with 340 M parameters (1.3 GB). We utilized the regular BERT model for transfer learning which is equipped with a maximum training size of 512 for each word embedding. BERT due to its bi-directional capabilities is able to hold context in its 768 layers. There are different approaches to utilizing its outputs and creating word embeddings with some just taking the last 12 layers, or concatenating the last four, or summing up the last four, etc. For the purpose of our work we summed up the last 12 layers since this provided the highest prediction accuracy. In order to create a document embedding we averaged all the word embeddings that belong to each document. This resulted to a matrix of size $N \times 768$.

\subsection{Classification}
As stated earlier, we used Python 3.8 and the main python package for our classifier was scikit-learn. We compared a variety of classifiers such as $K$-Nearest Neighbors (KNN) that incorporates distance metrics between its neighbors to decide which segment it belongs to, Support Vector Machine (SVM) that incorporates a hyper plane to define the distance between segments , and RandomForest (RF) a tree based algorithm that uses hundreds of decision trees to create a majority vote. We decided to use a RF model as our classifier due to its best prediction performance. We split the data into training and testing 0.67 and 0.33 percent respectively. We used grid search cross-validation to tune the hyper parameters of : criterion,  max depth of the trees, and number of estimator for each distinct document embedding dataset. We then took the best RF model and performed a stratified cross-Validation with a split of three to estimate the testing error. A stratified split of the data, ensures that we keep the same share of characterization, processing, modeling, and synthesis abstracts in each training and testing split in our cross-validation folds. We wanted to estimate the testing error before moving into explainability of the model to ensure we chose the best classifier from which to extract explanations.

\subsection{Local interpretability}
While accuracy metrics are a necessary and proper indicator of a classifier’s performance, practical uses of machine learning prove that the reasoning behind a machine’s decisions are just as necessary. Within the context of energetics research, we should have an understanding of why a topic label was predicted and which specific scientific terms informed this prediction for any research document. This makes the classification process reliable.

Interpretability is generally recognized on two scales, local and global, where global interpretability refers to methods that explain the whole model and local interpretability refers to methods that provide an explanation on an individual prediction \cite{rudin2021interpretable,linardatos2021explainable}. As we described above, within the context of the embedding experiments, we are more concerned with localized interpretations. We implemented LIME into our pipeline. LIME learns an interpretable model locally around the predictions that can explain the predictions of any classifier in an interpretable and faithful manner. This allows us, for any range of model complexity, to understand which terms were most important for each prediction. With the use of experts, we are able to confirm which model’s decision process aligns directly with how they would process and classify these documents. 

\section{Experimental results}\label{results}
We were able to obtain predictions and explanations for each of the models using the computing resources provided by the American University Zorro High Performance Computing System. With the intent of understanding the relationship between accuracy and interpretability for topic modeling on a scientific corpus, the evaluation process consisted of performance metrics and an expert’s ability to logically follow the output for the different embedding methods. 

\subsection{Prediction performance}
Four common classification metrics, accuracy, precision, recall and F1 score, were used to understand the classification performance of the different embedding methods. Accuracy measures how accurate our method is across all predictions. Continuing with precision, we are trying to gauge, that out of all our topic predictions, what percentage of the time were we rightly predicting topics. The more false positives we incur the lower the precision. For recall, we are measuring out of our predictions (true positives) of topic labels, how often were we actually correct? The more false negatives, the lower the recall. F1 is the harmonic mean balancing precision and recall. Our benchmark of performance is 0.25 since we are dealing with a four-label classification problem. Each model outperformed all the benchmark classification metrics by more than 50\%.

\begin{figure}[h]
\centering\includegraphics[width=11cm]{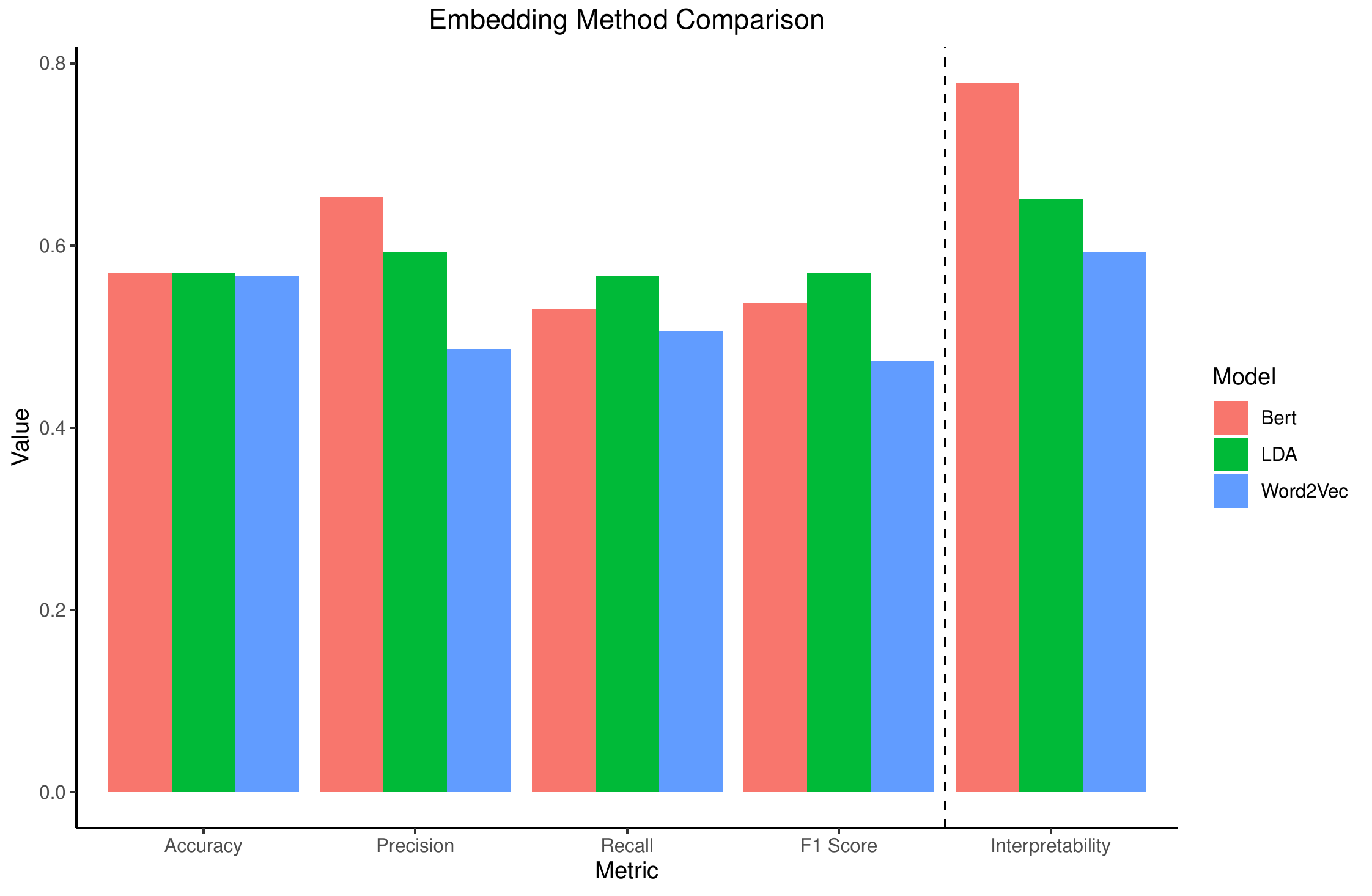} 
\centering
\caption{Prediction performance and agreement scores}
\label{result}
\centering
\end{figure}

From Figure \ref{result}, we see that BERT-based document embedding method outperformed the other three methods with regards to precision, however, LDA-based method outperformed with regards to F1 score and Recall. In terms of accuracy, the models performed similarly with a standard deviation of 0.0019. However, for recall the models showed greater variation, with a standard deviation among the models of 0.03. The standard deviation among the models for precision was 0.08, and with the BERT-based method performing best.

\subsection{Interpretability evaluation}
As discussed in previous section, in recent years, efforts have been made to extract information from a classifier. One such effort is the LIME system which produces local explanations for a classifier decisions. 
\begin{figure}[H]
\centering\includegraphics[width=18cm]{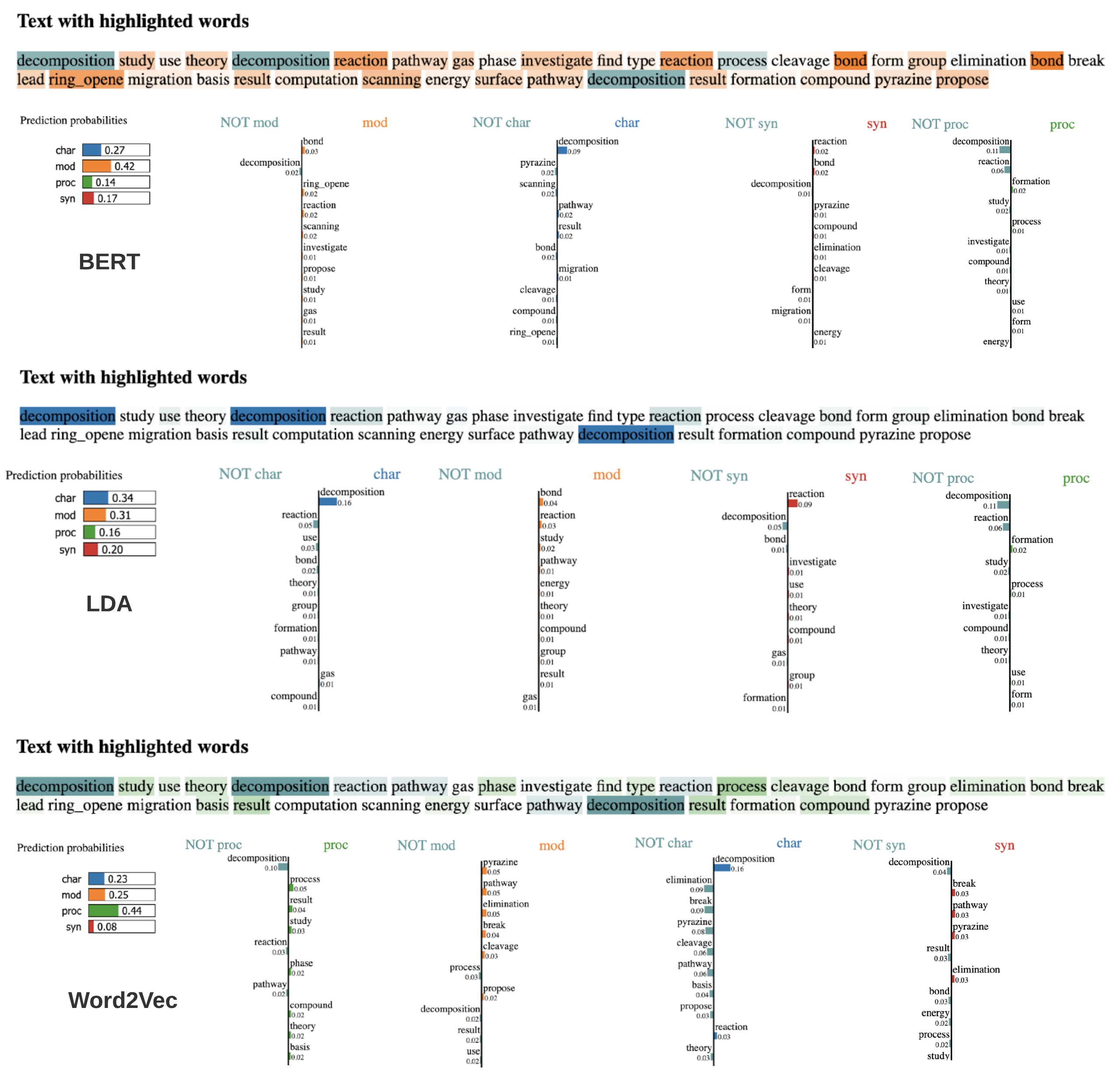} 
\centering
\caption{LIME output for a single testing observation which was correctly predicted as modeling by the BERT-based embedding method. Top: BERT Local explanations by LIME. Prediction probabilities are displayed in the upper left hand corner for each of the class labels. Furthermore, under each class label there are graphics showcasing how each word contributes positively or negatively toward each class label's prediction probabilities; Middle: LDA Local explanations by LIME; Bottom: Word2Vec Local explanations by LIME.}
\label{LIME}
\centering
\end{figure}
Figure \ref{LIME}, shows the LIME output for a single testing observation which was correctly predicted as modeling by the BERT-based embedding method. In the upper left hand corner, Figure \ref{LIME} displays the prediction probabilities for each of the class labels for all three embedding methods. In addition, under each class label there are graphics showcasing how each word contributes positively or negatively toward each class label's prediction probabilities. For instance, the top part in Figure \ref{LIME}, shows that our classifier using the BERT-based embedding method correctly predicted the target topic label as modeling with 0.42 percent probability. The word ``bond" contributes to 0.03 percent of that probability followed by the word ``ring opene"\footnote{It is worth mentioning that ``ring opene" is the pre-processed word for ``ring opening".}, and the word ``reaction". On the other hand, the middle part in Figure \ref{LIME}, shows that the LDA-based embedding method, for the same observation, provides a 0.31 probability percent of assigning the modeling topic but slightly favored the characterization label as its max probability at 0.34. For this case the word ``decomposition" contributed 0.16 percent towards that probability. Last, the last part in Figure \ref{LIME}, shows that the Word2Vec-based embedding method, for the same observation, provides a 0.44 probability percent of assigning the processing label and the word ``process" contributed 0.05 percent towards this probability. 

To quantitatively assess the interpretability of each embedding method, one of the subject matter experts in our team visually inspected and analyzed 86 LIME testing observation outputs and defined an agreement score. The agreement score is given by the fraction $c/N_{\rm{test}}$, where $N_{\rm{test}}$ is the number of testing abstracts and $c$ is the number of times a subject matter expert agreed with the LIME output using the following process:

\begin{itemize}
    \item[(1)] Collect the most significant terms as determined by LIME for the predicted label of a document.
    \item[(2)] Determine if any terms are too vague. If so, do they help tell a story when used with the surrounding terms? If not, the prediction output is illogical.
    \item[(3)] Determine if the terms relate better to another label. If so, the prediction is illogical.
    \item[(4)] If no terms are too vague and the story these terms tell relate to the predicted topic, the prediction is logical.
\end{itemize}

Based on this metric, we see from Figure \ref{result} that the BERT-based embedding method significantly outperformed all other methods in terms of interpretability. The experts were able to logically follow the LIME results and agree with the predictions over 78 percent of the time compared to 65 percent of the time with the LIME outputs using the LDA embeddings and 59 percent of the time with the Word2Vec embedding LIME outputs. 

\section{Conclusion}\label{conclusion}

The success of this study raises several interesting questions that can be explored in future work. Although the assumption that BERT-based embeddings carry meaningful characteristics and would yield reliable performance in terms of prediction accuracy and explainability, quantitatively evaluating such an assumption is a non-trivial task since BERT is mainly an unsupervised process. Thus, as a future task we propose to create formal settings where humans can evaluate whether a set of extracted embeddings have human-identifiable semantic coherence and where humans can evaluate whether the associations between a pre-labeled particular abstracts and a set of document embeddings make sense. These quantitative methods have been similarly used for measuring semantic meaning in inferred topics \cite{chang2009reading}. By developing human-based evaluation metrics, we will not only assess the document embedding space, but more importantly, we will be able to identify potential biases related to certain characteristics of the collected abstracts enabling us to correct our model before it is deployed at scale. In addition, comparing BERT with other popular latent variable methods as presented in \cite{moroney2021case}, would be of high interest. Finally, in terms of a computational chemistry perspective, the development of validation techniques for the extracted document embeddings and how they can be used for the discovery of energetic materials and systems is a significant research direction that deserves further investigation.

\subsubsection*{Acknowledgments}
Support for this work is gratefully acknowledged from the U.S. Office of Naval Research under contract number N00014-21-C-1016 and from the Energetics Technology Center under project number 2054.001.002.
\bibliographystyle{splncs04}
 
\bibliography{ref}

\end{document}